\documentclass[final,5p,times,twocolumn,number]{elsarticle}
\usepackage[switch]{lineno}
\usepackage{textgreek}
\usepackage[utf8]{inputenc}
\usepackage{graphicx}

\usepackage[table]{xcolor}

\usepackage{xurl}
\usepackage{amssymb}
\usepackage{amsmath,amsfonts,amsthm,bm}
\usepackage{bigints}
\usepackage{subcaption}
\usepackage{pgfplots}
\usepackage{comment}
\usepackage{enumitem}
\pgfplotsset{compat=1.7}
\usepgfplotslibrary{colorbrewer}
\usepgfplotslibrary{colormaps}
\usepackage{pifont}
\usepackage{booktabs}
\graphicspath{ {./Figures/} }

\usepackage{gensymb}
\usepackage{booktabs}

\usepackage{float}

\usepackage{siunitx}
\usepackage{titlesec}
\titleformat{\paragraph}
  {\normalfont\normalsize\bfseries}{\theparagraph}{1em}{}
\titlespacing*{\paragraph}{0pt}{3pt}{3pt}

\usepackage{setspace}
\biboptions{sort&compress}

\usepackage{caption}

\journal{Advanced Engineering Informatics}

\makeatletter
\def\ps@pprintTitle{%
 \let\@oddhead\@empty
 \let\@evenhead\@empty
 \def\@oddfoot{}%
 \let\@evenfoot\@oddfoot}
\makeatother

\usepackage{threeparttable}

\usepackage{hyperref}
\hypersetup{colorlinks=true, citecolor=red, linkcolor=blue, filecolor=blue, urlcolor=red,}

\newif\ifshowrevisions
\showrevisionsfalse   % hide colored revisions

\newcommand{\rev}[1]{%
  \ifshowrevisions
    \textcolor{blue}{#1}%
  \else
    #1%
  \fi
}

\begin{document}

\begin{frontmatter}

%===============================================================================
%	Title
%===============================================================================

\title {A Latency-Aware Framework for Visuomotor Policy Learning on Industrial Robots}

%============================================================================
%	Authors
%============================================================================

\author[add1]{Daniel Ruan\corref{cor1}}
\ead{daniel.ruan@princeton.edu}

\author[add1]{Salma Mozaffari\corref{cor1}}
\ead{salma.mozaffari@princeton.edu}

\author[add1]{Sigrid Adriaenssens}
\ead{sadriaen@princeton.edu}

\author[add1]{Arash Adel\corref{cor2}}
\ead{arash.adel@princeton.edu}

\address[add1]{Princeton University, Princeton, NJ 08544, USA}
\cortext[cor1]{Equal contribution.}
\cortext[cor2]{Corresponding author.}

%=========================================================================
%	Abstract
%=========================================================================

% Answer these questions for abstract:
% 1) What is the general problem addressed in the paper?
% 2) What is the specific research question to be answered?
% 3) What are the means and methods used by the authors to answer the stated question?
% 4) What is the answer to the research question?
% 5) Why is the answer important and for whom specifically?
% 6) How does the answer inspire future research

\begin{abstract}
% max 250 words (Advanced Eng. Inf. + Manufacturing systems requirement)

\rev{Industrial robots are increasingly deployed in construction and manufacturing tasks, where the deployment of end-to-end visuomotor policies is challenged by the observation–execution gap induced by observation, inference, and execution latencies. This gap is often significant on industrial robotic arms due to high-level control interfaces and slower closed-loop dynamics, making execution timing a dominant system-level concern. This paper presents a system-level, latency-aware framework for deploying and evaluating visuomotor policies on industrial robotic arms. The framework integrates latency-calibrated multimodal sensing, data synchronization, a unified communication pipeline, and a teleoperation interface for collecting expert demonstrations. Within this framework, we formalize a latency-aware execution strategy that assigns timestamps to policy-predicted action sequences and schedules only temporally feasible actions according to their intended execution time, enabling asynchronous inference and execution without modifying policy architectures or training procedures. We evaluate the framework on a contact-rich assembly task while systematically varying inference latency and compare latency-aware execution against blocking and naive asynchronous baselines using identical policies and sensing modalities. Results show that latency-aware execution preserves smooth motion, compliant contact behavior, and task progression consistent with demonstrations across inference latencies of 100--500 ms. Latency-aware execution maintained task duration and motion smoothness within 13\% and 9\% of the demonstration reference, respectively, while avoiding the latency-dependent slowdown observed under blocking execution and the large contact-force overshoots produced by naive asynchronous execution. These findings demonstrate that explicit handling of the observation--execution gap is essential for reproducible, closed-loop deployment of visuomotor policies on industrial robotic platforms.}

\end{abstract}

%===============================================================================
%	Keywords
%===============================================================================
% max 7 words (for AdvEngInf)

\begin{keyword}\small

Latency-aware execution, Industrial robots, Construction robotics, Visuomotor policy learning, Contact-rich robotic assembly, VR teleoperation
\end{keyword}

%=======================================================================
\end{frontmatter}

%=======================================================================
%                             INTRODUCTION
%=======================================================================
\section{Introduction}\label{introduction}

Construction and manufacturing systems increasingly rely on robotic automation to improve efficiency, repeatability, and worker safety in complex manipulation and assembly tasks. In construction in particular, persistent challenges such as slow productivity growth, labor shortages, and high injury rates motivate the adoption of robotic systems~\cite{delgado2019, wei2023, laukkanen1999, arndt2005}. To address these challenges, industrial robotic arms have been increasingly deployed in construction due to their high payload capacity and long reach, offering potential gains in both safety and productivity. However, unlike highly structured manufacturing settings, construction environments are characterized by substantial uncertainty, including fabrication inaccuracies, material variability, and dynamic on-site conditions~\cite{adel2024, chen2025}. These sources of uncertainty fundamentally undermine the effectiveness of open-loop or semi-structured robot programming approaches commonly used in Architecture, Engineering, and Construction (AEC) research, particularly for dexterous or contact-rich manipulation tasks where small geometric or material deviations can lead to failure~\cite{apolinarska2021}. Addressing these limitations requires robotic systems capable of closed-loop, sensory-driven execution, motivating the use of control- and learning-based methods that can continuously adapt robot actions during task execution.

Learning-based visuomotor control policies offer a promising approach to enabling such adaptive behavior, having demonstrated high success in dexterous manipulation by directly mapping multimodal sensory observations to robot actions~\cite{finn2017, levine2018, brohan2023}. Recent advances in robot learning, including diffusion-based policies and generalist robot manipulation policies such as vision-language-action models (VLAs), further highlight the potential of visuomotor policy learning to scale across tasks with complex dynamics and long horizon execution~\cite{chi2024a, kim2024}. At the same time, recent research has begun to explicitly recognize the observation--execution gap, induced by observation latency, policy inference latency, and policy execution latency, as a fundamental challenge for visuomotor control, motivating latency-aware execution strategies~\cite{liao2025, black2025, tri2026}. While many existing systems implicitly mitigate this gap through action chunking or asynchronous execution, these approaches typically rely on tightly integrated, low-latency control stacks and relatively fast system dynamics, in which inference latency is small relative to the control cycle~\cite{chi2024b, zhao2023}. 

The industrial robotic arms commonly used in construction operate under observation--execution gaps that are substantially larger than those in high-frequency, low-latency tabletop platforms. They typically expose only safety-certified, high-level control interfaces with limited update rates, while exhibiting slower closed-loop response due to high inertia, long-reach kinematics, nonlinear coupled dynamics, and conservative controller gains imposed by safety constraints~\cite{kommey2025, delgado2022}. \rev{As a result, the observation--execution gap becomes a dominant system-level challenge rather than a secondary implementation concern. By increasing sensitivity to stale actions, this gap makes precise temporal alignment among observation, inference, and execution essential for smooth, reliable, and reproducible learning-based workflows.} While learning-based methods have been demonstrated on industrial robotic arms~\cite{liu2022, chu2025}, they generally assume tightly integrated, low-latency execution pipelines and therefore do not explicitly treat the observation--execution gap as a core system-level consideration.

\begin{figure*}[!h]
  \centering
   \includegraphics[width=1\linewidth]{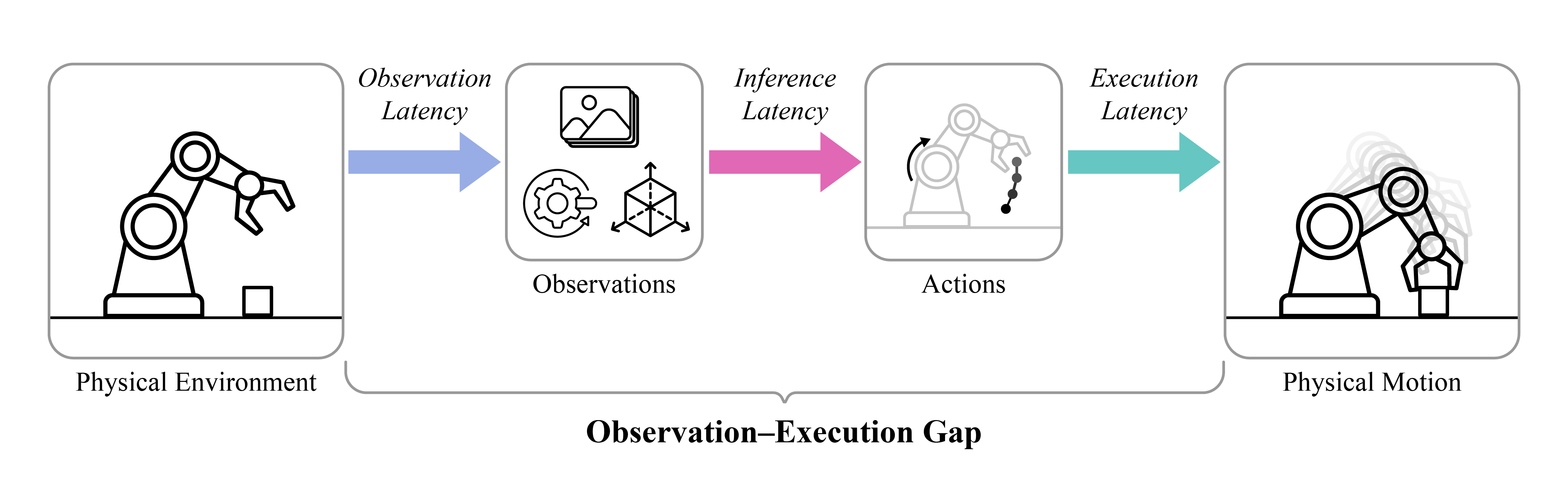}
   \caption{The observation--execution gap with non-negligible latency sources: observation, inference, and execution (see Section~\ref{latency_sources} for details).}
   \label{fig:latencies}
\end{figure*}

%%%%%%%%%%%%%%%%%%%%%%%%%%%%%%%%%%%%%%%%%%%%%%%%%%%%%%%%%%%%%%
% \subsection{Objectives and contributions}

This paper presents a system-level, latency-aware framework for deploying and evaluating visuomotor policies on industrial robotic arms operating under a non-negligible observation--execution gap induced by observation, inference, and execution latency (Fig.~\ref{fig:latencies}). The objective is to enable controlled and reproducible evaluation of closed-loop execution behavior by explicitly accounting for this gap at the system level. Rather than proposing new learning methods, the framework supports structured analysis of execution strategies under realistic industrial timing conditions.

The main contributions of this work are as follows:

\begin{itemize}[nosep]
    \item A reproducible, system-level experimental framework that integrates latency-calibrated multimodal sensing, temporally consistent data synchronization, and a unified communication and control pipeline for executing visuomotor policies on industrial robotic arms.
    \item A latency-aware execution strategy within this framework that formalizes time-aligned scheduling of policy-predicted action sequences, enabling asynchronous inference and execution without modifying policy architectures or training procedures.
    \item A controlled experimental study on a contact-rich construction assembly task that isolates the impact of execution strategy on closed-loop motion and interaction behavior under varying inference latency on industrial robotic hardware.
\end{itemize}

%========================================================================
%                             RELATED WORK
%========================================================================
\section{Related work}\label{sec:related_work}

In this section, we first discuss robotic assembly in construction, where industrial robotic arms are widely employed, and which forms the application context for our study. We then review visuomotor policy learning methods, focusing on how observation, inference, and execution latency are handled in existing systems. Finally, we examine prior work on robot learning with industrial robotic arms, highlighting system-level constraints that distinguish these platforms from high-frequency, low-latency tabletop robots.

\subsection{Robotic assembly in construction} 

Construction environments pose particularly demanding requirements for robotic assembly, spanning a diverse range of robotic platforms, from industrial robotic arms repurposed from factories to collaborative and mobile robots~\cite{chen2025}. Compared to factory manufacturing, robotic assembly in construction is characterized by inherent sources of uncertainty, such as fabrication inaccuracies (e.g., deviations in grasp pose, robot pose, and material processing), material imperfections (e.g., dimensional deviations, deformations, and warping), and dynamic site conditions, including occlusions and environmental variability~\cite{gandia2022, adel2024}. When these factors are not explicitly accounted for, robots executing open-loop or preprogrammed trajectories are prone to geometric imprecision, unintended contact, and task failure. These challenges are further amplified in dexterous or contact-rich manipulation tasks, where complex geometries, high contact forces, and tight tolerances make assembly highly sensitive to minor deviations~\cite{apolinarska2021}. As a result, many AEC workflows pair preprogrammed robot motions with human intervention to compensate for deviations and enable recovery from failure~\cite{adel2018, adel2020, helmreich2022, wang2024, skevaki2026, adel2022}. To reduce reliance on manual intervention, prior work has explored control-based methods or feedback-driven frameworks~\cite{dorfler2018, adel2024, cote2024, adel2025} and learning-based methods~\cite{apolinarska2021, yu2024, duan2024, duan2025, sun2026, mozaffari2026}. Compared with model-based control methods that rely on manually parameterized models and task-specific tuning, learning-based methods offer an alternative that can implicitly capture complex interaction dynamics, making them particularly attractive for closed-loop execution in domains characterized by high uncertainty and variability.

In contact-rich assembly, delayed or poorly timed actions can directly translate into force overshoot, jamming, or unstable interaction, making execution timing a critical system-level concern rather than a secondary control detail. These effects are exacerbated when robotic systems are accessed through high-level, buffered control interfaces, as is typical for industrial robotic arms. Since industrial robotic arms play a central role in large-scale robotic assembly in construction due to their high payload capacity and long reach~\cite{apolinarska2018, adel2020, lauer2023}, the following sections motivate a closer examination of the challenges that arise when applying modern robot learning methods to these systems.

%%%%%%%%%%%%%%%%%%%%%%%%%%%%%%%%%%%%%%%%%%%%%%%%%%%%%%%%%%%%%%%
\subsection{Visuomotor policy learning}

Visuomotor policy learning methods~\cite{levine2016, finn2017, levine2018, kalashnikov2018, brohan2023} have been increasingly adopted as alternatives to classical model-based control, owing to their strong performance in dexterous manipulation, including tasks involving deformable objects, contact-rich interactions, and bimanual coordination~\cite{zhao2023, zhao2025, chi2024a}. These policies map sensory--motor observations (e.g., proprioception, images, and force/tactile feedback) directly to robot actions. Among these, behavior cloning has emerged as a practical and widely used approach, replacing task-specific programming with human demonstrations, typically collected via teleoperation~\cite{seo2023, wang2023, shaw2023}. Building on these methods, recent advances have extended to generalist robot manipulation policies (also referred to as visuomotor foundation models), such as VLAs and large behavior models (LBMs), which aim to improve generalization, scalability, and robustness across highly dexterous tasks and unseen environments~\cite{kim2024, octomodel2024, tri2026, pi2025, gr2025a}. These models leverage large, heterogeneous datasets of real and synthetic demonstrations to learn broad task repertoires across varied environments and embodiments.

The effectiveness of visuomotor policy learning hinges not only on high-quality multimodal data and sensor fusion but also on properly accounting for the latencies induced by observation, inference, and execution~\cite{chi2024b, tri2026, liao2025, black2025}. Many existing methods implicitly mitigate these latencies through action chunking, receding-horizon execution, or asynchronous policy execution~\cite{chi2024a, zhao2023, chi2024b}. A complementary approach incorporates inference latency directly into policy training by conditioning action predictions on a fixed observation--action delay, thereby temporally aligning policy outputs with their execution without modifying the execution pipeline~\cite{liao2025}. While these strategies can improve temporal consistency in high-frequency control settings, they may still produce stale actions or out-of-distribution motion at chunk boundaries when inference latency is non-negligible or varies over time~\cite{black2025}.

Recent work on visuomotor foundation models has also begun to address inference and execution latency, motivated in part by the substantial computational cost of these models. For example, Black et al. treat asynchronous action chunking as an inpainting problem, freezing actions that are guaranteed to be executed while regenerating the remaining portion of the action sequence to incorporate newer observations~\cite{black2025}. Similarly, Liu et al. propose a test-time inference strategy that evaluates multiple candidate action sequences and selects those that balance temporal consistency and reactivity~\cite{liu2025}. Hierarchical VLA approaches further decompose control into a low-frequency planning module and a higher-frequency action generation module to manage inference latency~\cite{gr2025b, nvidia2025}. While these approaches can improve temporal alignment during execution, they typically introduce additional execution-time computation, which can be detrimental when deployed on industrial robotic systems, particularly in contact-rich tasks where delayed responses can lead to force overshoot or unstable interaction.

%%%%%%%%%%%%%%%%%%%%%%%%%%%%%%%%%%%%%%%%%%%%%%%%%%%%%%%%%
\subsection{Robot learning on industrial robots}

Many visuomotor policy learning methods, including those based on action chunking and asynchronous policy execution, are typically applied to tabletop robotic platforms with low-level, high-frequency control loops, fast system dynamics, and relatively direct access to motion commands, enabling reproducible execution under well-controlled timing assumptions~\cite{george2023, mete2024, chi2024a, zhao2025}. Industrial robotic arms, by contrast, are commonly accessed through safety-certified, high-level control interfaces with limited update rates, conservative controller gains, and restricted compliance, and they exhibit slower closed-loop response due to high inertia, long-reach kinematics, and nonlinear coupled dynamics~\cite{delgado2022, kommey2025}. These characteristics fundamentally shape how sensing and control are integrated in factory environments, where workflows often rely on offline programming, teach-pendant operation, or supervisory control rather than continuous closed-loop adaptation~\cite{zafar2024}. Under these conditions, the observation–execution latency becomes substantially more pronounced than in high-frequency systems, and delayed or stale actions can significantly degrade closed-loop performance. As a result, the observation--execution gap emerges as a dominant systems-level constraint, making precise temporal alignment among sensing, inference, and execution essential for smooth, reliable policy execution and for enabling reproducible learning-based workflows on industrial robotic platforms. 

A growing body of research has explored learning-based methods for industrial robotic arms in manufacturing and assembly tasks, including approaches that emphasize learning architectures, multimodal sensing, and human–robot interaction~\cite{liu2022, chu2024, wang2024, liu2024, chu2025}. However, many of these studies treat execution timing and control interfaces as fixed system properties, and often do not foreground the observation–execution gap induced by sensing, inference, and execution. As a result, while individual learning-based capabilities have been demonstrated on industrial robots, reproducible system-level frameworks that explicitly account for latency during closed-loop execution remain limited.

%========================================================================
%                             METHODS
%========================================================================
\section{Methodology}\label{methods}

This section describes the methodological foundations of the proposed latency-aware framework. We first introduce the abstractions, notation, and timing concepts required to reason about visuomotor policy execution under non-negligible sensing, inference, and execution latency (Section~\ref{preliminaries}). We then detail the experimental platform and system integration, including the industrial robotic hardware, communication stack, teleoperation interface for data collection, and sensing modalities used to instantiate the framework (Sections~\ref{setup} and~\ref{sensor_integration}). Finally, we present the latency-aware execution strategy that schedules policy-predicted actions according to their temporal feasibility under delayed inference and buffered control, along with the baseline execution strategies used for comparison (Section~\ref{execution_strategy}). Together, these components define a reproducible system-level pipeline for evaluating learning-based visuomotor policies on industrial robotic arms under realistic timing constraints.

%%%%%%%%%%%%%%%%%%%%%%%%%%%%%%%%%%%%%%%%%%%%%%%%%%%%%%%%%%%%%%%%%%%%%%%%%%
\subsection{Preliminaries}\label{preliminaries}

This section introduces the abstractions and notation used to describe the implementation and evaluation of visuomotor policies within our latency-aware framework on industrial robotic platforms. These definitions establish a common foundation for the experimental platform and execution strategy presented in the following sections.

\begin{figure*}[!t]
  \centering
   \includegraphics[width=0.8\linewidth]{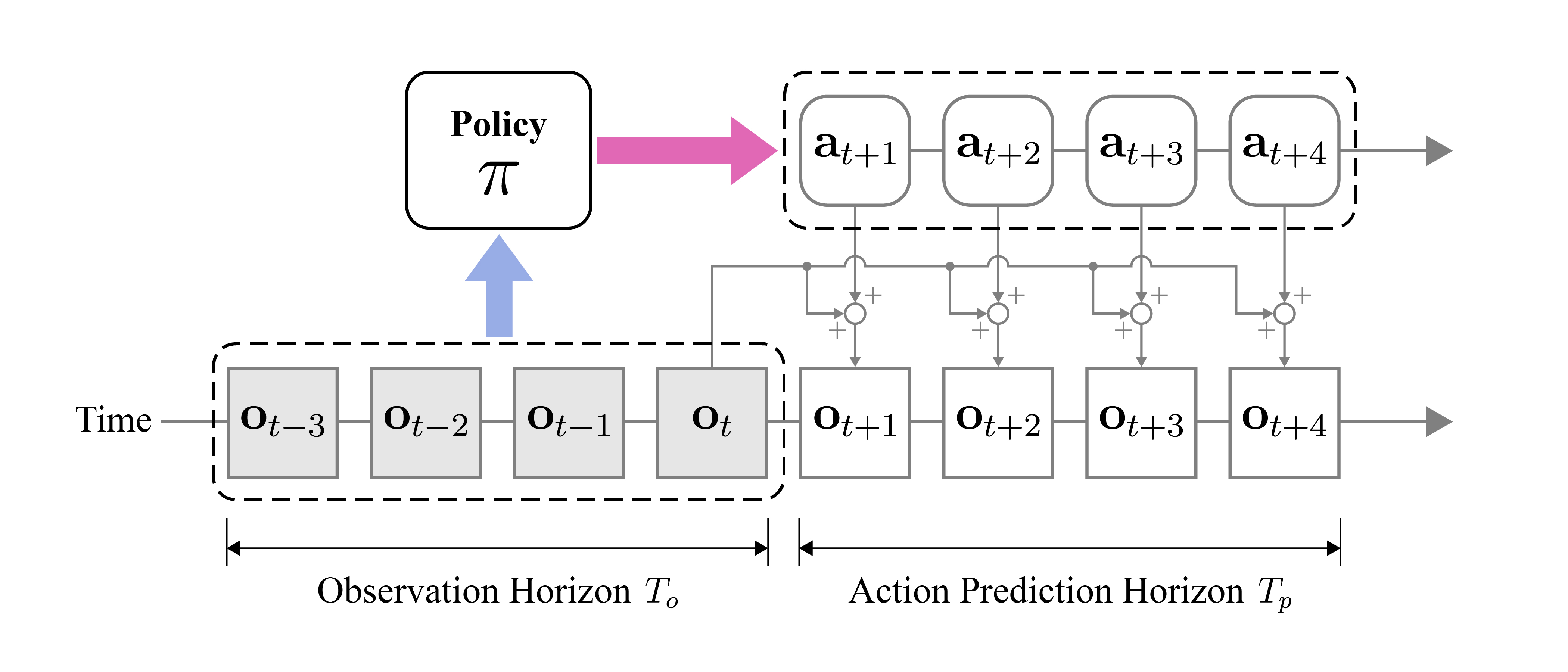}
   \caption{Policy interface.}
   \label{fig:policy_interface}
\end{figure*}

\subsubsection{Latency sources}\label{latency_sources}

Learning-based robotic systems deployed on physical hardware are subject to multiple sources of delay. We distinguish three latency components that jointly affect closed-loop behavior (Fig.~\ref{fig:latencies}):

\begin{itemize}[nosep]
    \item \emph{Observation latency} is the delay between a physical event at the robot or in the environment and the availability of the corresponding sensor measurement to the computing system. This includes sensor exposure time, onboard processing, communication delays, and middleware buffering.
    \item \emph{Inference latency} is the time required for a control policy to process observations and produce action prediction. This depends on the policy architecture, input dimensionality, and available computational resources.
    \item \emph{Execution latency} is the delay between issuing a command to the robot controller and the resulting physical motion. This includes communication delays, command buffering, internal controller dynamics, and actuator response.
\end{itemize}

These delays constitute the \emph{observation--execution gap}, defined as the temporal mismatch between the robot state represented by the observations used for inference and the robot state at the time when the corresponding actions are physically executed. If unaccounted for, this gap can lead to stale or poorly timed actions, resulting in idle behavior, overshoot, discontinuities, or unstable contact interactions~\cite{black2025, chi2024b}.

\subsubsection{Policy interface}\label{policy_interface}

Visuomotor policies map observations to robot actions, typically over short but finite time horizons~\cite{george2023, mete2024, chi2024a, zhao2025}. While specific architectures vary, many learning-based controllers share a common interface structure that is independent of the underlying learning method. \rev{We formalize this abstraction as the \emph{policy interface} used by the execution framework. The framework operates only on this input--output interface and does not depend on how the policy is represented or trained.} We assume the following characteristics (Fig.~\ref{fig:policy_interface}):

\begin{itemize}[nosep]
    \item Sensory observations are provided to the policy at discrete timesteps, even if underlying sensor streams operate in continuous time.
    \item At each inference step, the policy receives a fixed-length history of observations.
    \item The policy outputs a sequence of actions intended for execution over a finite future horizon, rather than an instantaneous control command.
    \item Actions are defined relative to the robot state represented in the observations used for inference.
\end{itemize}

Formally, let $T_o \geq 1$ denote the \emph{observation horizon} and $T_p \geq 1$ the \emph{action prediction horizon}. At a discrete timestep $t$, the policy $\pi$ receives an observation history $\mathbf{o}_{t-T_o+1:t}$ and produces a sequence of future actions:

\begin{equation}
\mathbf{a}_{t+1:t+T_p} = \pi\!\left(\mathbf{o}_{t-T_o+1:t}\right)
\end{equation}

\noindent where $\mathbf{a}_{t+i}$ denotes the action intended for execution $i$ timesteps after the observation at time $t$. \rev{This interface can represent both hand-specified and learned policies, including behavior cloning or diffusion-based visuomotor policies, provided that their outputs can be expressed as finite-horizon action sequences.}

\subsubsection{Observations and actions}\label{observations}

Sensors typically produce continuous-time measurements at modality-specific sampling rates. To construct discrete-time observations compatible with the policy interface, these streams must be resampled onto a common time grid. When future measurements are available, such as during offline demonstration processing, resampling can be performed using interpolation. When interpolation is infeasible or inappropriate, such as during policy evaluation or for modalities where interpolation is ill-defined (e.g., images), observations are formed using the most recent available measurements.

After resampling, each measurement is mapped to a vector representation through a modality-specific encoding function. These encoders may correspond to geometric parameterizations (e.g., poses or wrenches) or learned feature embeddings. Finally, the measurements for each timestep $t$ are concatenated into an observation vector $\mathbf{o}_t \in \mathbb{R}^n$. The resulting observation vector provides a unified representation for learning and policy inference, while preserving the task-relevant structure of each sensing modality.

During data processing for policy training, actions are computed from proprioceptive observations expressed in the pose representation space. Actions are defined as relative offsets with respect to the pose at the observation timestep. Given resampled poses $\mathbf{x} \in \mathbb{R}^d$, the action sequence associated with an observation at timestep $t$ is computed as:

\begin{equation}
\mathbf{a}_{t+i} = \mathbf{x}_{t+i} - \mathbf{x}_t,\quad i=1,...,T_p
\end{equation}

\subsubsection{Normalization}\label{normalization}

Observations and actions span different physical units and dynamic ranges, making normalization necessary for stable learning and consistent inference~\cite{tri2026}. All normalization parameters are computed from the resampled demonstrations and held fixed during policy evaluation. We adopt a quantile-based normalization scheme~\cite{pi2025, lee2025} that maps each scalar value $x$ to a normalized value $\hat{x} \in [-1,1]$:

\begin{equation}
\hat{x} = 2 \frac{x - q_{\ell}}{q_{u} - q_{\ell}} - 1
\end{equation}

\noindent where $q_{\ell}$ and $q_{u}$ denote lower and upper quantiles of the training distribution. The inverse mapping is applied during policy evaluation to recover physical units:

\begin{equation}
x = \frac{\hat{x} + 1}{2} (q_{u} - q_{\ell}) + q_{\ell}
\end{equation}

Normalization is applied independently to each observation and action dimension. This procedure reduces the influence of outliers while preserving the dynamic range of typical values and enables consistent treatment of heterogeneous sensing modalities within a single policy interface~\cite{tri2026}.

%moved here to show up in a proper location
\begin{figure*}[!h]
  \centering
   \includegraphics[width=1\linewidth]{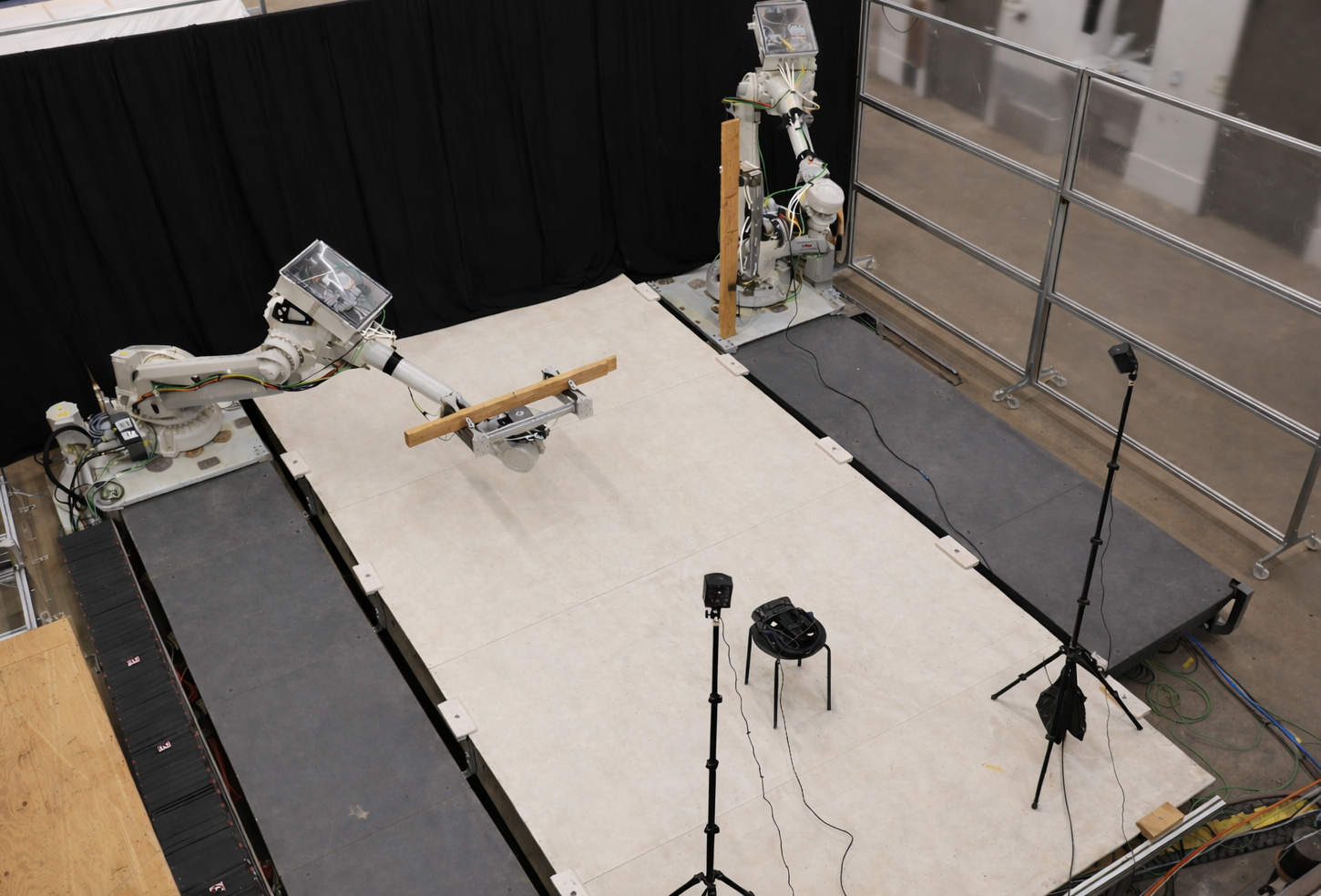}
   \caption{Experimental setup with two six-axis industrial robotic arms and a VR system used for teleoperation-based data collection.}
   \label{fig:setup}
\end{figure*}

%%%%%%%%%%%%%%%%%%%%%%%%%%%%%%%%%%%%%%%%%%%%%%%%%%%%%%%%%%%%%%%%%%%%%%%%
\subsection{Experimental platform}\label{setup}

This section describes the case study experimental platform used to evaluate the proposed latency-aware framework, including details of the communication stack and teleoperation interface. The experimental platform consists of an industrial robotic workcell comprising two industrial robotic arms, a robot controller, multiple programmable logic controllers (PLCs), a workstation for external computation, and a Virtual Reality (VR) system for teleoperation (Fig.~\ref{fig:setup}).

\begin{figure*}[!h]
  \centering
   \includegraphics[width=0.8\linewidth]{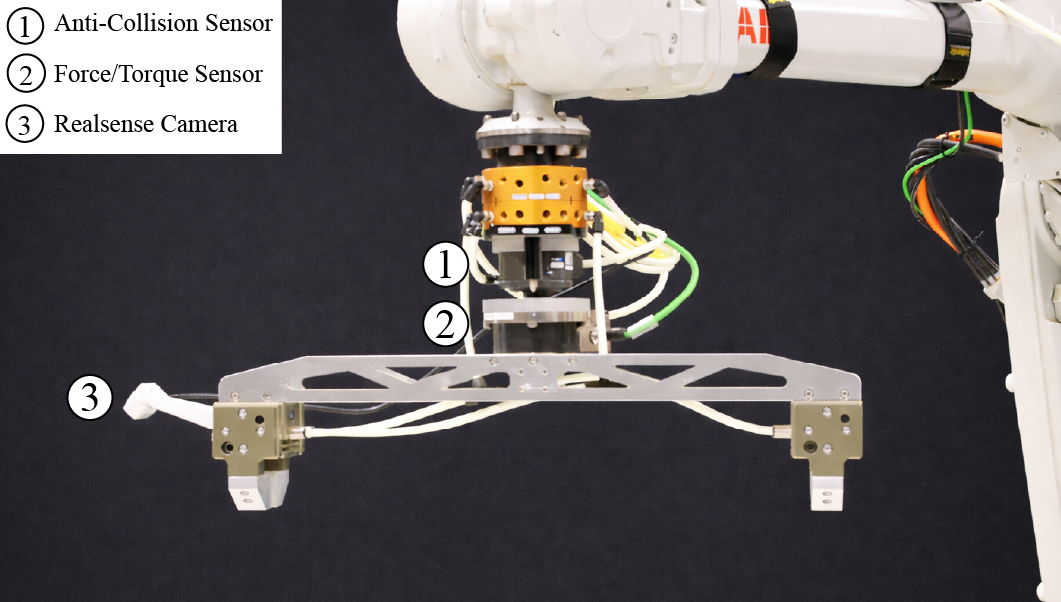}
   \caption{End effector with integrated force/torque sensor, anti-collision sensor, and eye-in-hand camera.}
   \label{fig:sensors}
\end{figure*}

The two industrial robotic arms are six-axis ABB manipulators\footnote{ABB IRB 4600~\cite{abb}}, each with a payload capacity of 40~kg and a reach of 2.55~m, mounted on linear tracks. The manipulators are equipped with custom pneumatically actuated end effectors designed to manipulate building-scale elements. One end effector integrates a six-axis force/torque sensor\footnote{ATI Delta IP60, with SI-330-30 calibration~\cite{ati}}, a pneumatic anti-collision sensor\footnote{Schunk OPR 081-P00~\cite{schunk-ac}}, and an eye-in-hand RGBD camera\footnote{Intel RealSense D435~\cite{realsense}} (Fig.~\ref{fig:sensors}). The anti-collision sensor provides passive compliance during contact and protects both the force/torque sensor and robot from excessive impact forces during execution.

The manipulators are controlled by an ABB IRC5 controller running RobotWare\footnote{RobotWare v6.16.01~\cite{robotware2024}} with the Externally Guided Motion (EGM) option~\cite{egm2021}. EGM exposes a Cartesian setpoint interface over a UDP-based connection, enabling an external device to stream target poses while the robot controller retains responsibility for low-level control, safety enforcement, and collision monitoring. Commands issued by external processes are therefore subject to communication delay, controller-side buffering, and internal execution latency. These characteristics make the platform well-suited to evaluating execution strategies under realistic industrial control constraints.

\subsubsection{Communication stack}\label{middleware}

The robot controller, PLCs, and workstation are interconnected through a dedicated local area network (LAN) that supports timestamped data exchange and command streaming. Peripheral devices, including gripper actuation and auxiliary sensors, are interfaced through local PLCs mounted on each arm that communicate with a central PLC via a fieldbus connection\footnote{TwinCAT~3~\cite{twincat2024}}. The central PLC aggregates signals from the robot controller and local PLCs and relays state and command streams between the robot and the workstation. \rev{In our implementation, the PLC network is used for low-bandwidth, timing-sensitive robot and peripheral signals, including robot/controller state, force/torque measurements, end-effector I/O, and anti-collision status signals. High-bandwidth sensing data, in particular the eye-in-hand RGB camera stream, are not routed through the PLC network, and instead acquired, processed, and synchronized directly on the workstation. This separation keeps high-bandwidth perception and inference outside the PLC communication path while preserving PLC-based routing for timing-sensitive control and safety signals.}

\begin{figure}[h]
  \centering
   \includegraphics[width=1\linewidth]{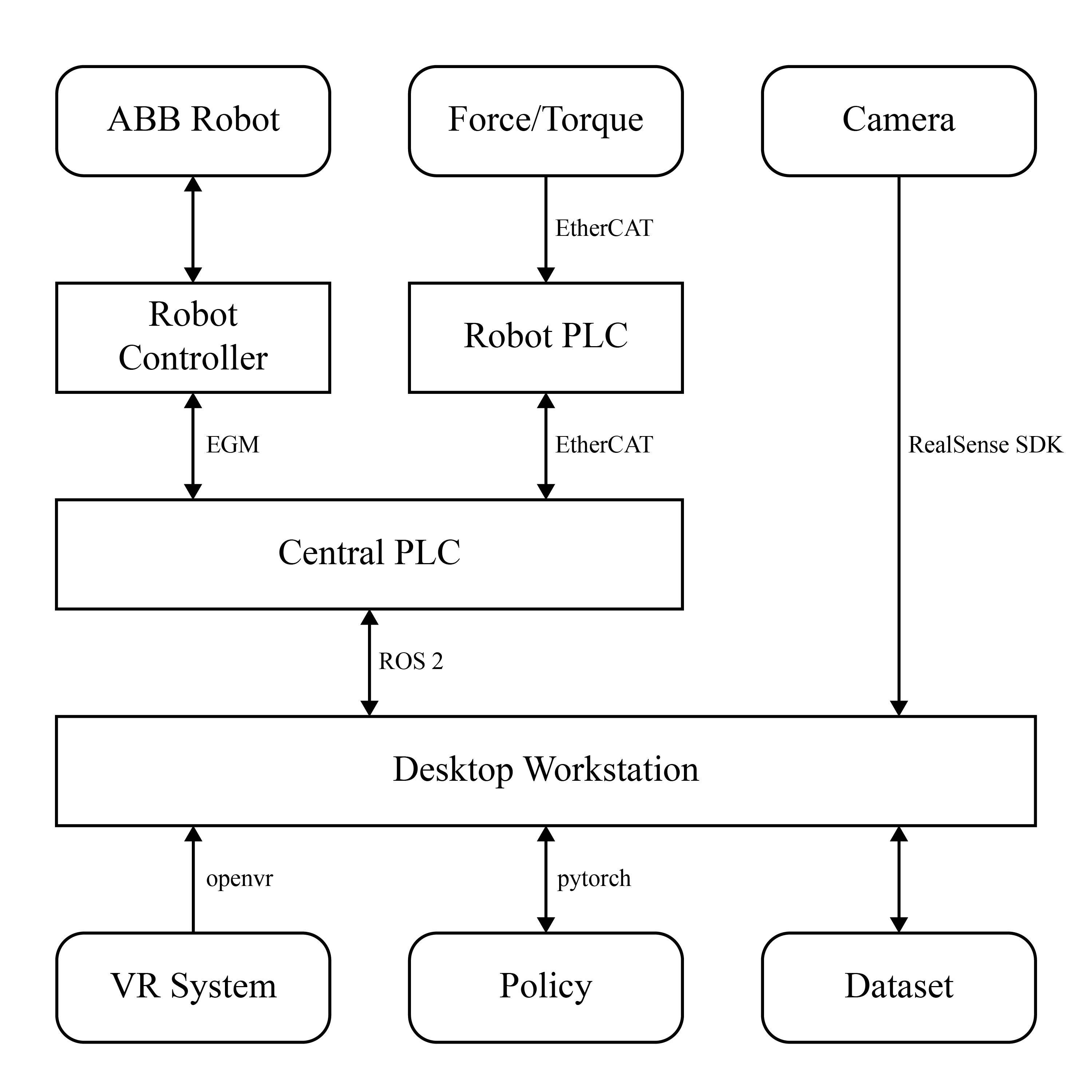}
   \caption{Hardware and software communication stack used in the experimental platform.}
   \label{fig:communication}
\end{figure}

Communication between system components is implemented using the Robot Operating System~2 (ROS~2)\footnote{ROS~2 Jazzy Jalisco~\cite{ros2}. To support Windows-based environments (e.g., the PLCs), ROS~2 is installed via RoboStack~\cite{Fischer2021}.}, which provides a distributed publish--subscribe middleware for streaming sensor measurements, robot state, and commands (Fig.~\ref{fig:communication}). \rev{All hardware devices and software components involved in the control loop are synchronized using Network Time Protocol (NTP)~\cite{Mills2010}. This provides a shared clock reference for time-stamped data exchange among the distributed control components, enabling consistent latency calibration across the industrial network.}

Observation latencies are calibrated offline and used to time-correct sensor measurements before publication (Section~\ref{sensor_integration}). \rev{Low-bandwidth latency-compensated state and peripheral streams} are aggregated within the PLC network and published by the central PLC to ROS~2 topics, which are subscribed to by the workstation for data collection and policy evaluation. Actions generated either by a policy or the teleoperation interface (Section~\ref{teleop}) are transmitted via ROS~2 and routed to the robot controller through a custom EGM driver. This design ensures that demonstration collection and policy evaluation share identical sensing, communication, and execution pathways, so observed differences in timing behavior can be attributed to inference latency and execution scheduling rather than to differences in system integration.

\subsubsection{Teleoperation interface}\label{teleop}

\begin{figure*}[!h]
  \centering
   \includegraphics[width=0.8\linewidth]{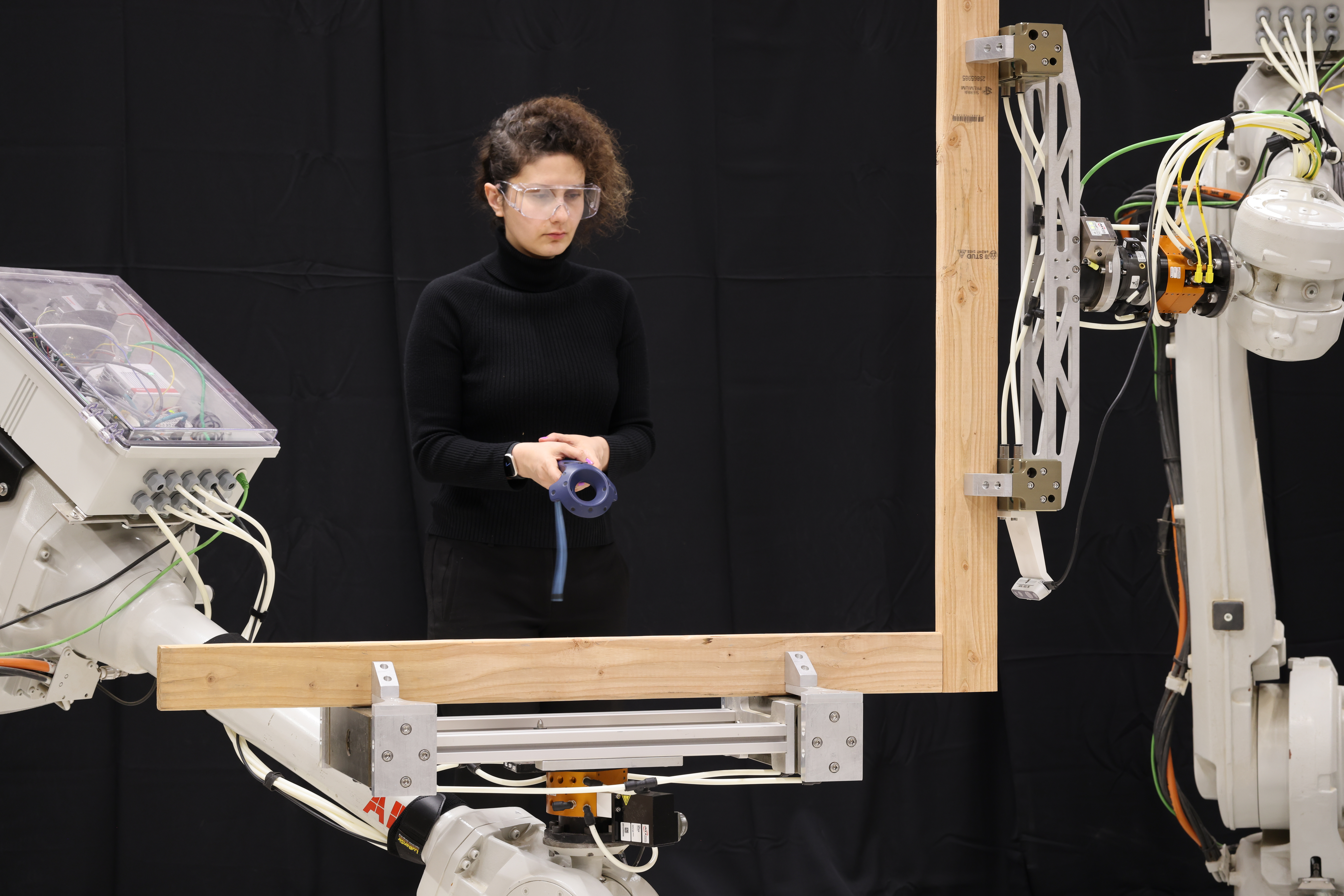}
   \caption{VR-based teleoperation system used for expert demonstration collection.}
   \label{fig:teleoperation}
\end{figure*}

Teleoperation is used to collect expert demonstrations under the same sensing, communication, and execution constraints used during policy evaluation. The goal of the teleoperation interface is not to optimize human performance, but to generate temporally consistent trajectories that are compatible with the latency-aware framework. Demonstrations are collected using a VR-based teleoperation system that maps human hand motion to robot end-effector motion in Cartesian space (Fig.~\ref{fig:teleoperation}). Similar VR-based teleoperation paradigms have been widely used for data collection for behavior cloning and human-in-the-loop control in robotic manipulation~\cite{seo2023, brohan2023, garg2021, luu2025}.

In our experimental platform, the human operator uses a VR hand controller\footnote{HTC VIVE Pro 2~\cite{htc}} to stream six-degree-of-freedom hand poses to the workstation via the OpenVR~\cite{openvr} application programming interface (API). These poses are converted to target robot poses and sent to the robot controller while the hand controller's trigger is held. When the operator first presses the trigger, the current hand pose and robot tool center point (TCP) pose are recorded as reference frames. Releasing the trigger pauses robot motion and reinitializes the reference frames upon re-engagement. This interaction pattern supports long-horizon demonstrations while limiting operator fatigue and reducing unintended drift.

While the trigger is held, hand controller motions are mapped into relative robot motion commands with respect to the stored reference frames. Rotational and translational components are treated separately and may be independently scaled to enable fine manipulation. Formally, teleoperation is implemented as a relative motion mapping. When the operator presses the controller trigger at time $t_0$, the hand controller pose $\mathbf{T}_c(t_0) \in \mathrm{SE}(3)$ and the robot tool center point (TCP) pose $\mathbf{T}_r(t_0) \in \mathrm{SE}(3)$ are recorded as reference frames. For subsequent times $t \ge t_0$, hand motion is interpreted as a relative transformation with respect to the controller reference:

\begin{equation}
\Delta \mathbf{T}_c(t) = \mathbf{T}_c(t_0)^{-1}\mathbf{T}_c(t)
\end{equation}

Because the VR tracking frame is not aligned with the robot world frame, rotational and translational components of the relative motion are applied separately. Let $\Delta \mathbf{R}_c(t) \in \mathrm{SO}(3)$ and $\Delta \mathbf{p}_c(t) \in \mathbb{R}^3$ denote the rotation and translation components of $\Delta \mathbf{T}_c(t)$. The commanded robot TCP pose is computed as:

\begin{equation}
\mathbf{R}_r(t) = \mathbf{R}_r(t_0)\exp\!\left( \alpha \log\!\left(\Delta \mathbf{R}_c(t)\right) \right)
\end{equation}

\vspace{-10pt}

\begin{equation}
\mathbf{p}_r(t) = \mathbf{p}_r(t_0) + \beta\, \Delta \mathbf{p}_c(t)
\end{equation}

\noindent where $\alpha > 0$ and $\beta > 0$ are user-defined scaling factors for rotational and translational motion, respectively. Rotational updates are applied via composition on $\mathrm{SO}(3)$, while translational displacements are applied additively in a shared world-aligned basis. This formulation avoids the need for precise extrinsic calibration between the VR tracking frame and the robot world frame while maintaining intuitive motion correspondence and stable long-horizon teleoperation.

Timestamped observation streams are recorded throughout teleoperation. After data collection, the observations are synchronized (Section~\ref{observations}) and post-processed to remove idle segments (i.e., where the difference between adjacent observations is below a set threshold). This post-processing prevents the overrepresentation of unintended stationary actions in the demonstrations (e.g., due to pauses for hand controller readjustment).

%%%%%%%%%%%%%%%%%%%%%%%%%%%%%%%%%%%%%%%%%%%%%%%%%%%%%%%%%%%%%%%%%%%%%%%%%%
\subsection{Sensor integration}\label{sensor_integration}

The experimental platform integrates multiple sensing modalities to capture the robot's state and physical interaction during demonstrations and policy evaluation. We employ proprioception, force/torque sensing, and vision (Fig.~\ref{fig:sensors}), which reflect common modalities utilized in learning-based manipulation using industrial robotic arms~\cite{apolinarska2021,chu2025, wang2024, chu2024}. This section details how each of these modalities is integrated into our experimental platform and policy interface; these methods can be generalized to other sensing modalities and industrial robotic platforms. 

\subsubsection{Proprioception}\label{proprioception}

The robot kinematic state is obtained via the controller’s external feedback interface (EGM) at a nominal update rate of about 83~\si{\hertz} (12~\si{\milli\second} cycle time). Each feedback message includes a timestamped Cartesian pose of the TCP, consisting of position and orientation. TCP poses from the robot controller are represented as 7D vectors (3D position and 4D unit quaternion). To avoid discontinuities associated with quaternions, TCP poses are mapped to a 9D representation using a continuous 6D rotation parameterization~\cite{zhou2019}. This representation enables standard vector-space operations required for normalization, regression, and differencing. Actions are represented in the same 9D space as relative pose offsets with respect to the observed TCP pose. During execution, a target pose is obtained by adding the predicted action vector to the current 9D pose representation and converting the resulting rotation back to a valid element of $\mathrm{SO}(3)$ via Gram–Schmidt~\cite{zhou2019}. This projection step enforces orthonormality before converting the rotation back to a unit quaternion for transmission to the robot controller.

Execution latency is estimated offline using trajectory alignment. The robotic arm is commanded to follow a time-indexed, constant-velocity Cartesian trajectory, and timestamped pose feedback is recorded. Execution latency is estimated by fitting linear models to the commanded and tracked trajectories and computing the relative time shift. In our experimental platform, the execution latency was approximately 225~\si{\milli\second}. This estimate is used by the execution strategy to schedule action commands sufficiently in advance of their intended execution time (Section~\ref{execution_strategy}).

\subsubsection{Force/torque}

Interaction forces are measured using a six-axis force/torque sensor mounted at the robot wrist (Fig.~\ref{fig:sensors}), with an internal sampling rate of 3000~\si{\hertz}. The sensor applies an internal infinite impulse response (IIR) low-pass filter to attenuate high-frequency components, and the filtered measurements are resampled at 60~\si{\hertz} before publication. This filtering reduces measurement noise caused by mechanical vibration and limits high-frequency artifacts that are not relevant to the contact dynamics studied in this work. Measurements are expressed as spatial wrenches and timestamped at acquisition by a local PLC before being published to the ROS~2 network. The Force/torque (wrench) measurements are represented as 6D vectors expressed in the sensor frame $\{f\}$, denoted as ${}^{f}\bm{\mathcal{F}}_{\mathrm{F/T}} \in\mathbb{R}^6$.

To isolate interaction forces from configuration-dependent load effects, gravity compensation is applied to the measured wrench ${}^{f}\bm{\mathcal{F}}_{\mathrm{F/T}}$ to remove the contribution of the end effector and the grasped object. The resulting gravity-compensated wrench ${}^{f}\bm{\mathcal{F}}_{\mathrm{ext}}$ is used directly as an observation input to the policy. Explicit gravity compensation is necessary to ensure that force/torque signals used for learning and evaluation reflect contact dynamics rather than static weight-induced forces, which would otherwise dominate the measurements and confound execution-level analysis under delayed control. 

Gravity compensation is implemented using a standard rigid-body wrench formulation~\cite{lynch2017}. Let $M$ denote the combined mass of the end effector and grasped object. The gravitational force $\mathbf{f}_g \in \mathbb{R}^3$ expressed in the world frame $\{w\}$ is:

\begin{equation}
\mathbf{f}_g = [0,\,0,\,-Mg]^\top
\end{equation}

\noindent where $g$ denotes the gravitational constant. The associated gravitational wrench ${}^{w}\bm{\mathcal{F}}_g \in \mathbb{R}^6$ expressed in ${\{w\}}$, is then given by:

\begin{equation}
{}^{w}\bm{\mathcal{F}}_g= 
\begin{bmatrix}
\mathbf{f}_{g}\\
\bm{m}_{g}
\end{bmatrix}
\end{equation}

\noindent \noindent where $\bm{m}_{g} \in \mathbb{R}^3$ is the torque generated by $\mathbf{f}_{g}$ in the world frame ${\{w\}}$. Let ${}^{w}\mathbf{T}_{f} \in \mathrm{SE}(3)$ denote the pose of the force/torque sensor frame $\{f\}$ in the world frame, with rotation ${}^{w}\mathbf{R}_{f}$ and translation ${}^{w}\mathbf{p}_{f}$. Wrenches are transformed from the world frame to the sensor frame using the adjoint transformation $[\mathrm{Ad}_{{}^{w}\mathbf{T}_{f}}] \in \mathbb{R}^{6 \times 6}$:

\begin{equation}
[\mathrm{Ad}_{{}^{w}\mathbf{T}_{f}}]
=
\begin{bmatrix}
{}^{w}\mathbf{R}_{f} & [{}^{w}\mathbf{p}_f]{}^{w}\mathbf{R}_{f} \\
\mathbf{0} & {}^{w}\mathbf{R}_{f}
\end{bmatrix}
\end{equation}

\noindent where $[{}^{w}\mathbf{p}_f] \in \mathbb{R}^{3\times3}$ denotes the skew-symmetric matrix associated with the translation vector ${}^{w}\mathbf{p}_f$. The gravitational wrench ${}^{w}\bm{\mathcal{F}}_g$ expressed in the sensor frame is then:

\begin{equation}
{}^{f}\bm{\mathcal{F}}_g = [\mathrm{Ad}_{{}^{w}\mathbf{T}_{f}}]\,{}^{w}\bm{\mathcal{F}}_g
\end{equation}

\noindent and the gravity-compensated external wrench used by the policy is computed as:

\begin{equation}
{}^{f}\bm{\mathcal{F}}_{\mathrm{ext}} =
{}^{f}\bm{\mathcal{F}}_{\mathrm{F/T}} -
{}^{f}\bm{\mathcal{F}}_g
\end{equation}

The mass and center of mass of the gripping assembly are estimated from the CAD model under an assumption of homogeneous material distribution, with densities adjusted to match the measured physical mass.

\subsubsection{Vision}

Visual observations are acquired using an eye-in-hand RGB-D camera mounted on the end effector (Fig.~\ref{fig:sensors}). In this work, only the RGB stream is used. Images are captured at a resolution of $320 \times 240$ pixels and a nominal frame rate of 60~\si{\hertz}. \rev{Unlike the low-bandwidth robot state and force/torque streams, the image stream is not transmitted through the PLC network and instead is acquired directly by the workstation, where timestamp correction and image processing are performed before synchronization with other modalities.} To map images into a vector representation suitable for policy inference, we employ a fixed, non-learned visual encoding based on Histograms of Oriented Gradients (HOG)~\cite{dalal2005}. Images are resized to $96 \times 96$ pixels, converted to grayscale, and encoded using HOG with 6 orientation bins, $16 \times 16$ pixel cells, $2 \times 2$ cells per block, and L2-Hys block normalization, resulting in a 600D feature vector. HOG is chosen because it provides a compact, deterministic representation of local edge and shape structure while remaining computationally lightweight and free of learned parameters. This choice yields predictable, repeatable inference latency, which is essential for isolating the effects of execution scheduling and temporal alignment in our system-level evaluation (Section~\ref{experiments}), independent of the learned visual feature extractors.

Camera observation latency is calibrated offline using a timestamp-based method adapted from prior work~\cite{chi2024b}. A QR code encoding the system time is displayed on a monitor and captured by the camera, allowing the observation latency to be estimated as the difference between the encoded timestamp and the image reception time at the workstation. In our experimental platform, the camera observation latency was approximately 82~\si{\milli\second}.

%%%%%%%%%%%%%%%%%%%%%%%%%%%%%%%%%%%%%%%%%%%%%%%%%%%%%%%%%%%%%%%%%%%%%%%%%%
\subsection{Latency-aware execution}\label{execution_strategy}

\begin{figure*}[h]
  \centering
   \includegraphics[width=0.8\linewidth]{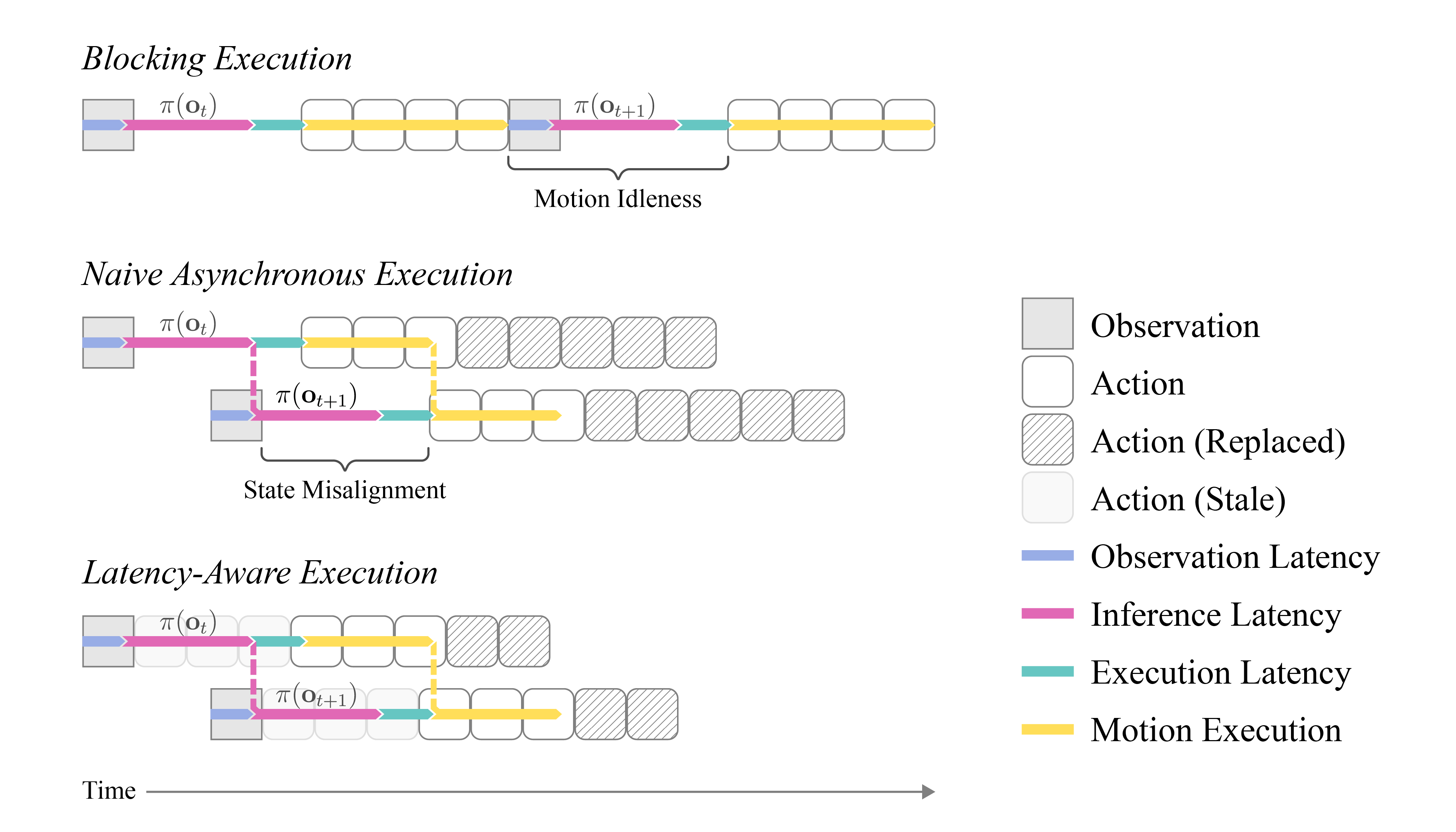}
   \caption{Comparison of the execution strategies. The latency-aware execution is our proposed method.}
   \label{fig:latency_matching}
\end{figure*}

This section details our execution strategy for deploying visuomotor policies on industrial robotic arms within our proposed latency-aware framework. We do not introduce a new execution algorithm. Instead, we systematize latency-aware, time-aligned action scheduling, as used in prior robot learning systems~\cite{chi2024b}, and integrate it into our unified latency-aware framework tailored to the constraints of industrial robotic arms.

The execution strategy schedules policy-predicted actions based on their feasibility under the observation--execution gap (Fig.~\ref{fig:latency_matching} bottom). Rather than modifying the policy architecture or training procedure, the strategy operates at the boundary between the policy interface and the industrial robot controller, explicitly accounting for execution latency when selecting commands to send. Policy outputs are treated as timestamped action sequences defined over a finite action prediction horizon $T_p$ (Section~\ref{policy_interface}). These actions are stored in a buffer indexed by their intended execution time. Inference runs asynchronously with respect to command execution and may update the buffer at irregular intervals.

When a policy is evaluated, the input observation is associated with a continuous-time timestamp $\tau_{\mathrm{obs}}$. The policy output $\{\mathbf{a}_{t+1}, \mathbf{a}_{t+2}, \dots, \mathbf{a}_{t+T_p}\}$, originally indexed in discrete timesteps, is converted into a continuous-time action sequence by assigning execution timestamps based on the nominal policy execution period $\Delta \tau$. Specifically, each predicted action $\mathbf{a}_{t+i}$ is assigned the timestamp:

\begin{equation}
\mathbf{a}(\tau_{\mathrm{obs}} + i\,\Delta \tau), \quad i = 1, \dots, T_p
\end{equation}

\noindent This conversion maps the policy’s discrete prediction horizon into a time-indexed action sequence aligned with the execution clock of the control interface. At each command update cycle, the execution module queries the current time and computes the target execution timestamp by adding the estimated execution latency of the robot controller (Section~\ref{proprioception}). The action corresponding to this target timestamp is then retrieved from the buffer. If the target timestamp lies between two buffered actions, the executed command is obtained via linear interpolation. \rev{Linear interpolation serves as a sufficient local approximation between predicted actions in our experiments because the interpolation interval is short (see Section~\ref{policy}). Higher-order interpolation or explicit trajectory smoothing may affect execution stability, but their effects depend on the policy rate, robot controller, and contact dynamics, and are therefore outside the scope of this paper.}

The resulting action is sent to the robot controller at the current update cycle. Actions in the buffer whose intended execution times precede the current target timestamp are considered stale and discarded. When a new policy inference result becomes available, it replaces the contents of the buffer with a newly timestamped action sequence. This receding-horizon update rule ensures that the buffer always reflects the most recent feasible future actions without retroactively altering commands that should already have been executed.

Formally, let $\tau_{\mathrm{now}}$ denote the current time at a command update cycle, and let $\delta$ denote the estimated execution latency of the robot controller (Section~\ref{proprioception}). The action selected for execution is the buffered action sampled at time $\tau = \tau_{\mathrm{now}} + \delta$. If no valid buffered action exists at or after $\tau$, the system issues a safe fallback command by holding the previously commanded pose. This strategy enables asynchronous inference and execution while maintaining temporal consistency between policy predictions and physical robot motion under non-negligible inference and execution latency.

\subsubsection{Baseline execution strategies}\label{baseline}

To contextualize the role of latency-aware action scheduling, we consider two commonly used execution strategies as baselines (Fig.~\ref{fig:latency_matching}):

\begin{itemize}[nosep]
\item \emph{Blocking execution}: policy inference and execution are serialized (also referred to as \emph{naive synchronous inference}~\cite{black2025}). Command updates are paused while inference is performed, after which a fixed number of predicted actions are executed. This strategy avoids executing stale actions but introduces idle time proportional to inference latency. 

\item \emph{Naive asynchronous execution}: policy inference and execution run concurrently, and predicted actions are streamed to the robot controller as soon as they become available, without compensating for inference delay or execution latency. This strategy minimizes idle time but may execute actions outside their intended temporal context.
\end{itemize}

%========================================================================
%                             EXPERIMENTS
%========================================================================
\section{Experimental design}\label{experiments}

This section evaluates how inference latency and execution scheduling affect closed-loop behavior when deploying a visuomotor policy on an industrial robotic arm within the proposed latency-aware framework. The objective is not to assess policy learning performance or generalization, but to isolate the impact of execution strategy under a non-negligible observation--execution gap. The evaluation focuses on how different execution strategies mediate the temporal alignment between policy predictions and physical robot motion. By deploying a deterministic policy for a contact-rich assembly task, observed differences in rollout behavior can be attributed primarily to execution scheduling rather than to \rev{policy-specific learning effects, inference stochasticity}, or task variability.

\subsection{Task description}\label{task}

\begin{figure*}[!h]
  \centering
   \includegraphics[width=\linewidth]{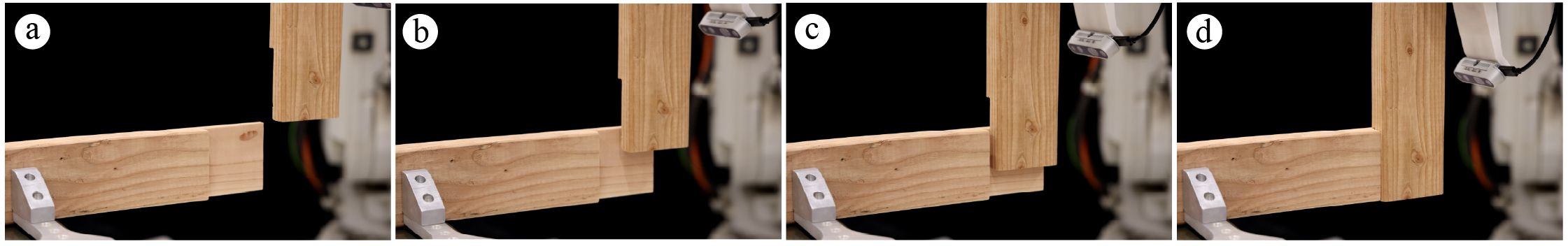}
   \caption{Corner lap connection task from start to end: (a) start point in free space, (b) initial contact, (c) sliding under contact, and (d) completed task.}
   \label{fig:task}
\end{figure*}

The experimental task is a contact-rich assembly of a timber corner lap joint (Fig.~\ref{fig:task}). One robotic arm inserts a vertical timber stud into a fixed horizontal stud with a matching slot geometry. So, the robot gripping the horizontal stud is not actively controlled, while only the robot with the vertical timber is commanded. The task proceeds through free-space motion, contact initiation, and sustained sliding under contact until the final assembled configuration is reached. Once contact is established, the geometry constrains motion along multiple axes, making execution behavior sensitive to timing misalignment and stale actions. This task is intentionally selected because free-space motion is relatively tolerant to latency, whereas contact onset and sliding phases are highly sensitive to delayed or poorly timed commands. As a result, differences in execution scheduling manifest not only in task duration but also in force profiles, motion smoothness, and interaction stability.

Twenty expert demonstrations are collected using the VR-based teleoperation interface described in Section~\ref{teleop}. The dataset is deliberately small and structured to avoid introducing robustness or generalization effects that could obscure execution-related behavior. Across demonstrations and policy evaluations, the initial pose of the manipulated vertical element is fixed, end-effector orientation is held constant, and control is restricted to translational TCP motion. Demonstrations and rollouts share identical sensing, synchronization, and execution pipelines, ensuring that demonstrations provide a meaningful reference for evaluating execution behavior under matched system conditions.

\rev{For the assembly task considered here, proprioceptive and force/torque observations may be sufficient for successful execution; the visual modality is included to demonstrate synchronization of heterogeneous sensing streams with distinct latency characteristics within the proposed framework. Modality ablations, varying initial conditions, rotational control, and external perturbations would provide useful insight into task-specific policy robustness, but are outside the scope of the present evaluation.}

\subsection{Policy instantiation}\label{policy}

To eliminate variability from policy training and stochastic inference, we employ a deterministic $k$-nearest neighbors (k-NN) policy as a proxy for a learned visuomotor controller. This choice ensures that policy outputs are fully determined by the demonstration dataset and the distance metric, allowing execution-related effects to be isolated.

The policy operates with an observation horizon $T_o = 1$, an action prediction horizon $T_p = 16$, and $k = 5$. These values match the assumptions of the policy interface described in Section~\ref{policy_interface}. Each observation vector comprises a 9D end-effector pose representation, a 6D gravity-compensated force/torque wrench, and a 600D visual feature vector (Section~\ref{sensor_integration}), resulting in a 615D observation space.

After latency calibration, all sensing streams are synchronized to a common discrete-time grid at 10~\si{\hertz} (i.e., $\Delta \tau=$ 100 ms). This rate is chosen as a practical and commonly used synchronization frequency for quasi-static tasks in prior visuomotor learning work~\cite{chi2024a}. For offline demonstration processing, \rev{synchronization to this grid is achieved by interpolating proprioceptive and force/torque streams and by assigning each timestep the most recent available visual observation.} During policy evaluation, all modalities use the most recent available measurements to respect causal constraints. This distinction ensures that training data are temporally consistent without introducing non-causal assumptions during execution.

Observations and actions are normalized using the quantile-based scheme described in Section~\ref{normalization}, with $q_{\ell}=0.01$ and $q_u=0.99$. Because the k-NN policy relies on a fixed distance metric, feature scaling directly determines the relative influence of each sensing modality. To prevent high-dimensional visual features from dominating nearest-neighbor retrieval, normalized observation dimensions are rescaled by modality-specific factors proportional to their dimensionality. More specifically, all feature dimensions belonging to modality $m$ are scaled by $\sqrt{1 / d_m}$, where $d_m$ is the modality dimensionality.

\begin{table*}[!h]
    \centering
    \caption{Comparison of median evaluation metrics between the demonstration reference (Ref.), Blocking Execution (BE), Naive Asynchronous Execution (NAE), and Latency-Aware Execution (LAE) strategies at various inference latencies (IL). For each inference latency, bolded values indicate the metric closest to the demonstration reference. \rev{The corresponding rollout-level variability is visualized by the box plots in Fig.~\ref{fig:metric_boxplots}}.}
    \label{tab:metrics_summary}
    \begin{tabular}{
        l
        S[table-format=3.2]
        |S[table-format=3.2] S[table-format=4.2] S[table-format=2.2]
        |S[table-format=2.2] S[table-format=3.2] S[table-format=2.2]
        |S[table-format=2.2] S[table-format=4.2] S[table-format=2.2]
    }
        \toprule
        & \multicolumn{1}{c|}{}
        & \multicolumn{3}{c|}{100 ms IL}
        & \multicolumn{3}{c|}{300 ms IL}
        & \multicolumn{3}{c}{500 ms IL} \\
        \cmidrule(lr){3-5}
        \cmidrule(lr){6-8}
        \cmidrule(lr){9-11}
        Evaluation metric
        &
        \multicolumn{1}{c|}{Ref.}
        & \multicolumn{1}{c}{BE}
        & \multicolumn{1}{c}{NAE}
        & \multicolumn{1}{c|}{LAE}
        & \multicolumn{1}{c}{BE}
        & \multicolumn{1}{c}{NAE}
        & \multicolumn{1}{c|}{LAE}
        & \multicolumn{1}{c}{BE}
        & \multicolumn{1}{c}{NAE}
        & \multicolumn{1}{c}{LAE} \\
        \midrule
        Task duration (s)
        & 13.60
        & 18.36 &  9.33 & \bfseries 12.08
        & 20.72 &  9.03 & \bfseries 12.18
        & 23.64 &  9.93 & \bfseries 11.90 \\

        Idle ratio (\%)
        & 5.10
        & 23.40 &  1.17 &  \bfseries 1.42
        & 37.53 &  \bfseries 4.98 &  4.10
        & 43.78 &  \bfseries 4.48 &  5.80 \\

        Contact force (N)
        & 73.98
        & 231.71 & 319.71 &  \bfseries 37.99
        &  \bfseries 56.02 & 180.43 &  35.41
        &  42.72 & 329.32 &  \bfseries 45.28 \\

        \rev{Peak force (N)}
        & \rev{154.65}
        & \rev{819.89} & \rev{829.85} & \rev{\bfseries 140.10}
        & \rev{\bfseries 204.69} & \rev{795.09} & \rev{65.50}
        & \rev{\bfseries 163.01} & \rev{851.38} & \rev{64.73} \\

        Force smoothness (N/s)
        & 136.06
        & 807.67 & 1229.10 &  \bfseries 41.58
        &  \bfseries 54.84 &  918.86 &  31.12
        &  \bfseries 33.15 & 1153.26 &  33.03 \\

        Motion smoothness (m/s$^{3}$)
        & 0.47
        & 0.85 & 0.61 & \bfseries 0.47
        & 0.82 & 0.70 & \bfseries 0.51
        & 0.70 & 0.66 & \bfseries 0.46 \\
        \bottomrule
    \end{tabular}
\end{table*}

\subsection{Experimental conditions}\label{experimental_conditions}

We evaluate our latency-aware execution strategy against two baseline execution strategies (Section~\ref{baseline}). All three are implemented within the same experimental platform, sharing identical sensing, synchronization, communication, and control interfaces. This ensures that observed differences in execution behavior arise from execution scheduling rather than from differences in system integration.

\begin{itemize}[nosep]
    \item \textbf{Blocking execution (baseline)}: policy inference and execution are serialized, causing command updates to pause while inference is performed.
    \item \textbf{Naive asynchronous execution (baseline)}: policy inference and execution run concurrently, and predicted actions are streamed to the robot controller as soon as they become available, without compensating for inference or execution latency.
    \item \textbf{Latency-aware execution}: policy-predicted actions are explicitly scheduled according to their temporal feasibility under the estimated observation--execution gap, as described in Section~\ref{execution_strategy}.
\end{itemize}

Inference latency is artificially varied by inserting fixed delays of 100~\si{\milli\second}, 300~\si{\milli\second}, and 500~\si{\milli\second} into the inference loop. These values span a representative range of latencies encountered in modern visuomotor policies\rev{: on the workstation used in this study, Diffusion Policy~\cite{chi2024a} with 10 DDIM inference steps takes $49 \pm 1$~\si{\milli\second} per inference call, while the standard 100-step DDPM variant takes $466 \pm 9$~\si{\milli\second}}. We execute and record twenty rollouts for each combination of execution strategy and inference latency.

\subsection{Evaluation metrics}\label{evaluation_metrics}

Execution behavior is evaluated using metrics that capture temporal efficiency, motion smoothness, and interaction quality. \rev{All twenty rollouts successfully complete the task across each execution strategy and inference latency condition ($180/180$ rollouts in total). Because binary task success does not differentiate between execution strategies in this setting, evaluation focuses on execution quality metrics that capture how the task is completed.} The following metrics are computed for each rollout and compared against the demonstration reference:

\begin{itemize}[nosep]
    \item \textbf{Task duration}: total time from motion onset to task completion.
    \item \textbf{Idle ratio}: fraction of execution time during which the end effector is effectively stationary, defined as periods where the Cartesian TCP velocity magnitude is below 1~mm/s.
    \item \textbf{Contact force}: root mean square (RMS) of the measured force magnitude during contact phases, defined as time intervals where the force magnitude exceeds 5~N.
    \rev{\item \textbf{Peak force}: Maximum instantaneous force magnitude recorded during contact phases, capturing transient force overshoot.}
    \item \textbf{Force smoothness}: RMS of the time derivative of the measured force magnitude, capturing high-frequency force fluctuations and oscillatory interaction behavior.
    \item \textbf{Motion smoothness}: RMS of Cartesian jerk, penalizing start--stop behavior and abrupt velocity changes.
\end{itemize}

For a time-varying scalar signal $x(t)$ evaluated over a time interval $\mathcal{T}$, the RMS value is computed as:

\begin{equation}
\mathrm{RMS}(x) = \sqrt{\frac{1}{|\mathcal{T}|} \int_{\mathcal{T}} x(t)^2 \, \mathrm{d}t},
\end{equation}

\noindent where $|\mathcal{T}|$ denotes the duration of the interval. For metrics defined over subsets of the execution (e.g., contact phases), the integral is evaluated only over the corresponding time intervals. In practice, all RMS-based metrics are computed using time-weighted integration to account for nonuniform sampling. The same metrics are computed on demonstration trajectories to provide a consistent reference for comparison.

\rev{For each metric and latency setting, we compared latency-aware execution with each of the baseline strategies using two-sided Mann-Whitney $U$ tests on the 20 rollout values per condition~\cite{Mann1947}. $p$-values are Bonferroni-corrected by multiplying by the number of comparisons performed per metric--latency family~\cite{Dunn1961}.}

%moved here to show up at a proper location in text
\begin{figure}[!h]
    \centering
    \includegraphics[width=1\linewidth]{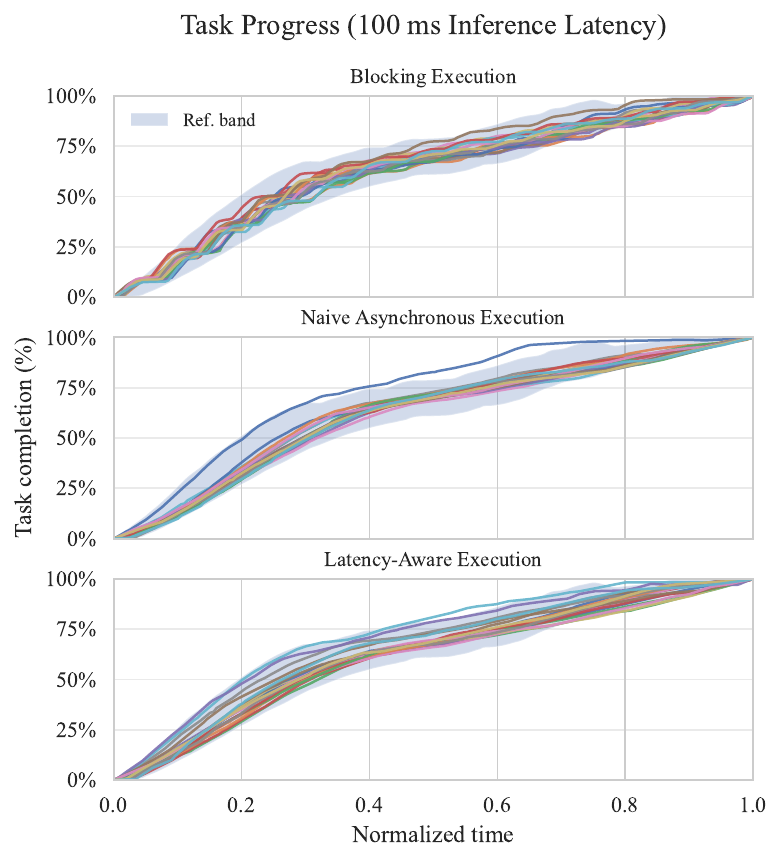}
    \caption{Task progression for each execution mode at 100~\si{\milli\second} inference latency, shown as normalized distance to the goal over execution time. The shaded region indicates the distribution of demonstration trajectories, while colored lines correspond to individual rollouts.}
    \label{fig:task_progress_100}
\end{figure}

%========================================================================
%                             RESULTS
%========================================================================
\section{Results and discussion}\label{results}

\begin{figure}[!h]
    \centering
    \includegraphics[width=1\linewidth]{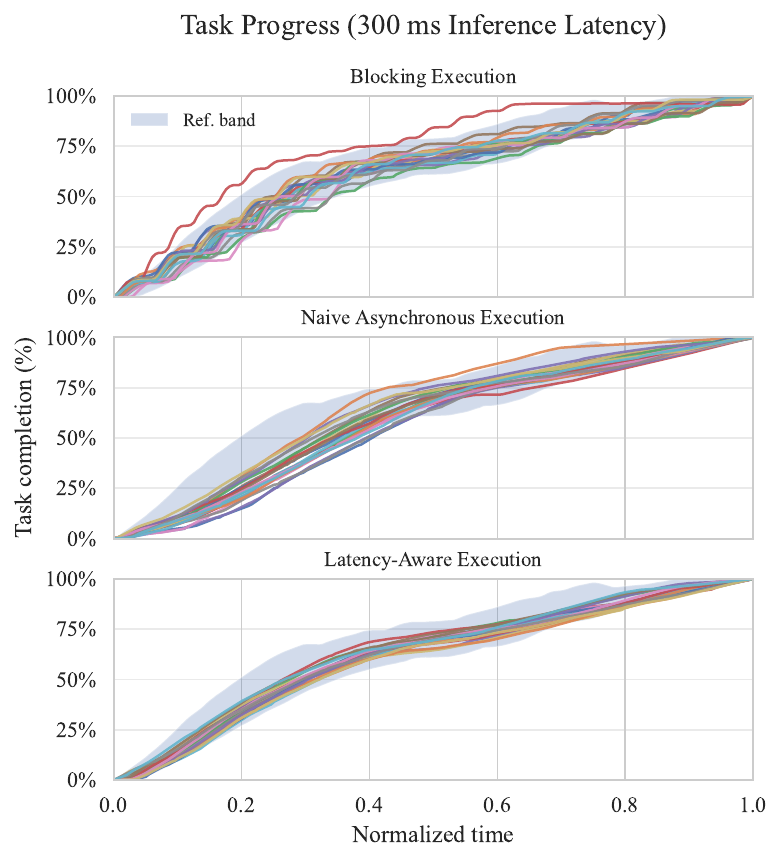}
    \caption{Task progression for each execution mode at 300~\si{\milli\second} inference latency, shown as normalized distance to the goal over execution time. The shaded region indicates the distribution of demonstration trajectories, while colored lines correspond to individual rollouts.}
    \label{fig:task_progress_300}
\end{figure}

\begin{figure}[!h]
    \centering
    \includegraphics[width=1\linewidth]{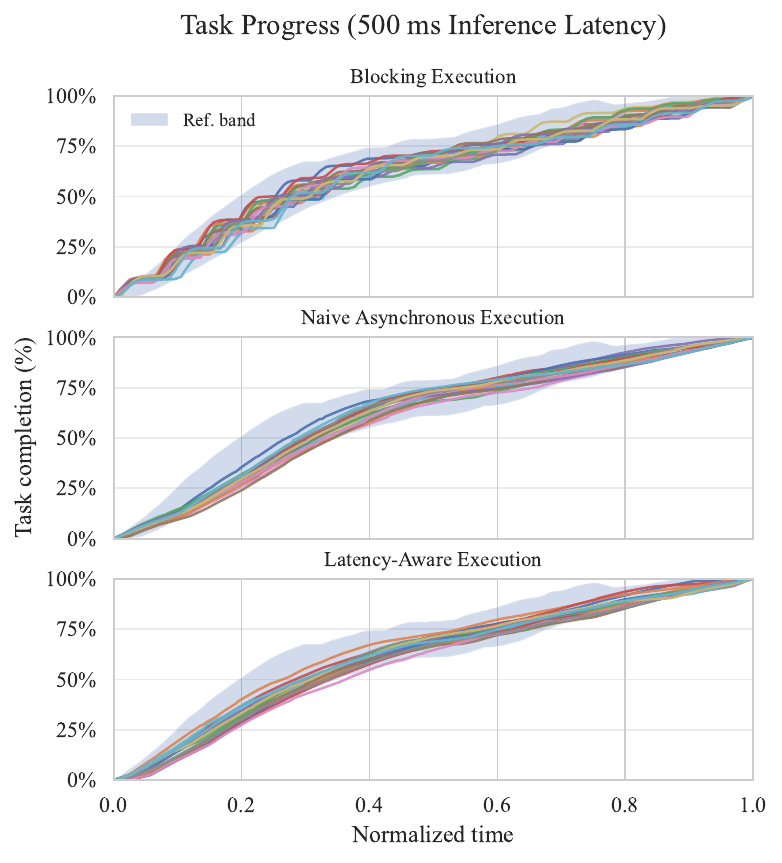}
    \caption{Task progression for each execution mode at 500~\si{\milli\second} inference latency, shown as normalized distance to the goal over execution time. The shaded region indicates the distribution of demonstration trajectories, while colored lines correspond to individual rollouts.}
    \label{fig:task_progress_500}
\end{figure}

Across all inference latencies, execution strategy consistently dominates closed-loop behavior. Latency-aware execution maintains task progression trajectories close to the demonstration reference at 100~\si{\milli\second}, 300~\si{\milli\second}, and 500~\si{\milli\second} inference latency (Figs.~\ref{fig:task_progress_100}--\ref{fig:task_progress_500}), whereas blocking and naive asynchronous execution exhibit qualitatively different deviations from the demonstration reference. Quantitative summaries of all evaluation metrics are reported in Table~\ref{tab:metrics_summary}.

\begin{figure*}[p]
    \centering

    \begin{subfigure}[t]{0.48\textwidth}
        \centering
        \includegraphics[width=\linewidth]{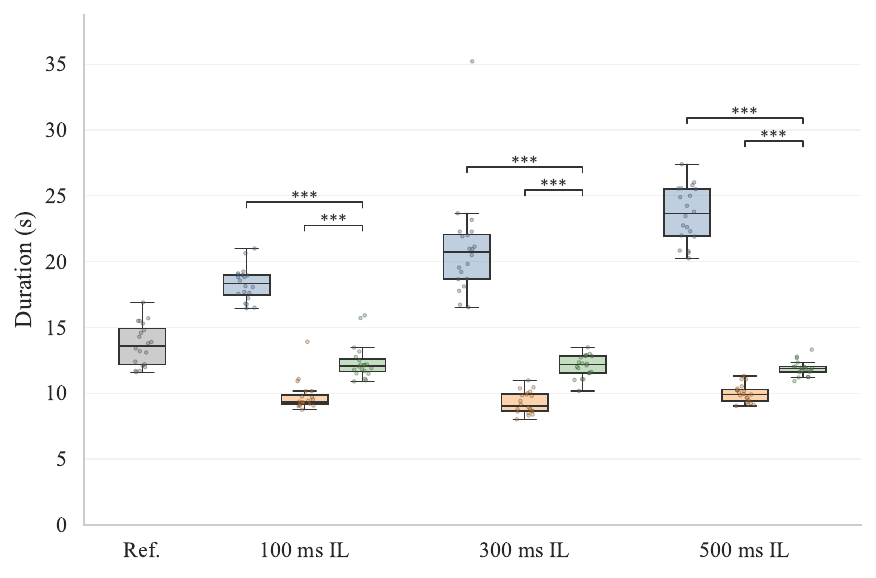}
        \caption{Task duration}
        \label{fig:task_duration}
    \end{subfigure}
    \hfill
    \begin{subfigure}[t]{0.48\textwidth}
        \centering
        \includegraphics[width=\linewidth]{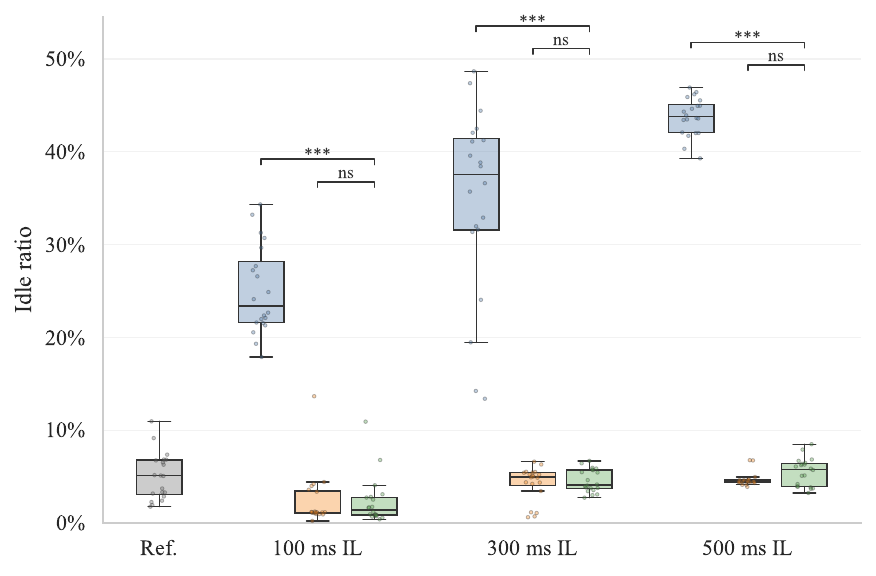}
        \caption{Idle ratio}
        \label{fig:idle_ratio}
    \end{subfigure}

    \vspace{0.75em}

    \begin{subfigure}[t]{0.48\textwidth}
        \centering
        \includegraphics[width=\linewidth]{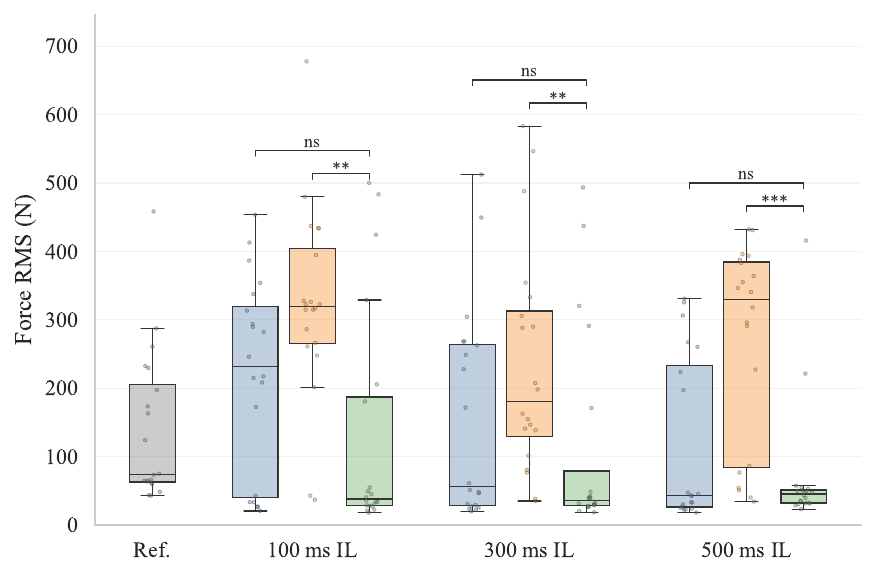}
        \caption{Contact force}
        \label{fig:contact_force}
    \end{subfigure}
    \hfill
    \begin{subfigure}[t]{0.48\textwidth}
        \centering
        \includegraphics[width=\linewidth]{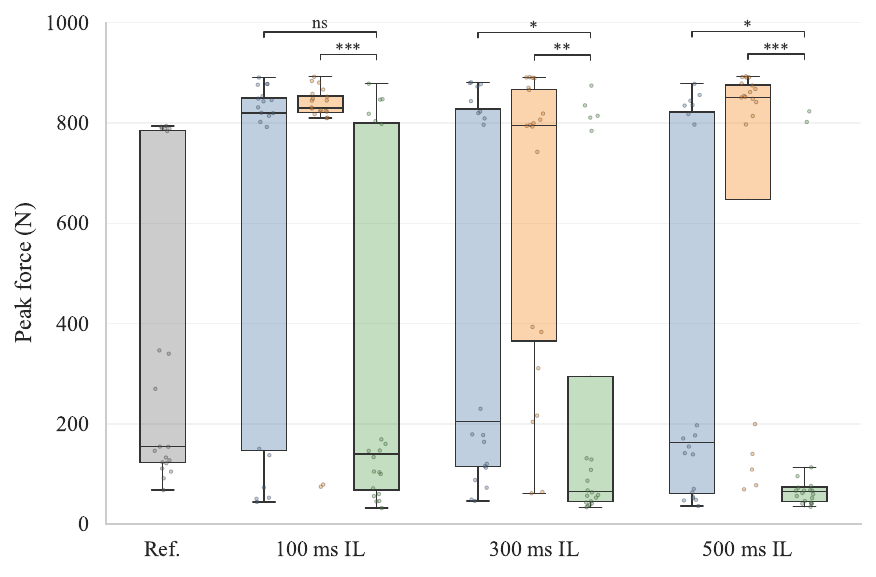}
        \caption{Peak force}
        \label{fig:peak_force}
    \end{subfigure}

    \vspace{0.75em}

    \begin{subfigure}[t]{0.48\textwidth}
        \centering
        \includegraphics[width=\linewidth]{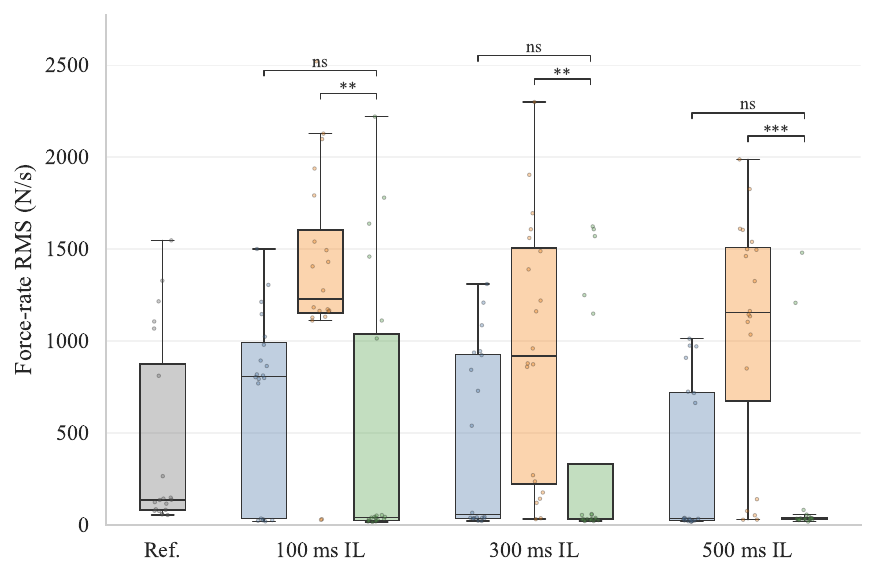}
        \caption{Force smoothness}
        \label{fig:force_smoothness}
    \end{subfigure}
    \hfill
    \begin{subfigure}[t]{0.48\textwidth}
        \centering
        \includegraphics[width=\linewidth]{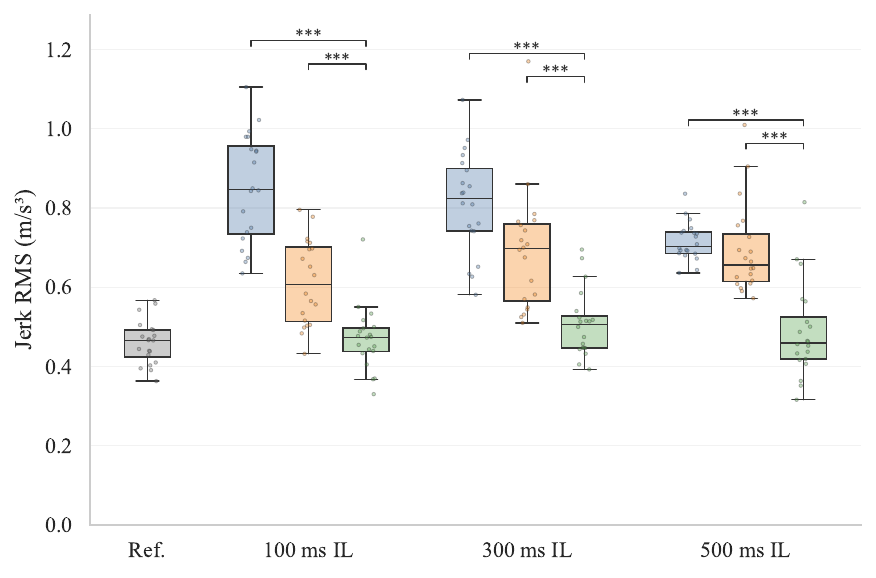}
        \caption{Motion smoothness}
        \label{fig:motion_smoothness}
    \end{subfigure}
    
    \includegraphics[width=0.3\textwidth]{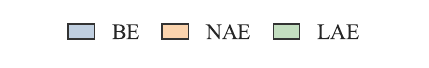}

    \caption{\rev{Comparison of the evaluation metrics between the demonstration reference (Ref.), Blocking Execution (BE), Naive Asynchronous Execution (NAE), and Latency-Aware Execution (LAE) strategies at inference latencies (IL) of 100, 300, and 500~\si{\milli\second}. Boxes span the interquartile range (IQR) with the median shown as a horizontal line; whiskers extend to 1.5$\times$IQR. Individual rollout values are overlaid as scatter points. Brackets indicate pairwise differences between LAE and each baseline strategy assessed by two-sided Mann-Whitney $U$ tests with Bonferroni correction: ns (not significant); $^{*}$ ($p<0.05$); $^{**}$ ($p<0.01$); and $^{***}$ ($p<0.001$).}}
    \label{fig:metric_boxplots}
\end{figure*}

Blocking execution preserves temporal correctness by avoiding the execution of stale actions but incurs increasing idle time as inference latency grows. This effect is reflected in longer task durations and high idle ratios, particularly at 300~\si{\milli\second} and 500~\si{\milli\second} latency (Figs.~\ref{fig:task_duration},~\ref{fig:idle_ratio}). The repeated pauses introduced by serialized inference and execution also lead to degraded motion smoothness due to start--stop behavior (Fig.~\ref{fig:motion_smoothness}).

Naive asynchronous execution minimizes idle time and yields the shortest task durations across all latency settings (Figs.~\ref{fig:task_duration},~\ref{fig:idle_ratio}). However, it frequently departs from the demonstration reference, exhibiting \rev{force overshoot} and increased force variability. These effects are especially pronounced during contact initiation and sliding, where delayed or poorly timed actions lead to force spikes and oscillatory interaction behavior (Figs.~\ref{fig:contact_force}--\ref{fig:force_smoothness}). \rev{In particular, naive asynchronous execution produced median peak forces roughly 5.1--5.5~times that of the demonstration reference across all latency levels, while overall contact forces reached 2.4--4.5~times.}

\rev{Latency-aware execution reflects a trade-off between responsiveness and contact stability rather than uniformly optimizing every metric. Latency-aware execution balances the effects of the two baseline execution strategies by preserving asynchronous execution while scheduling only temporally feasible actions, yielding stable task durations and contact forces closer to the demonstration reference across latency settings.} \rev{Across all inference latencies, motion smoothness remained within 9\% of the demonstration reference  (Fig.~\ref{fig:motion_smoothness}), and task duration remained within 13\% of the demonstration reference, contrasting with the latency-dependent increase observed under blocking execution.} Task progression under latency-aware execution also closely follows the demonstration distribution across all latency settings (Figs.~\ref{fig:task_progress_100}--\ref{fig:task_progress_500}).

The impact of execution scheduling is most pronounced during contact phases, where executing stale or poorly timed actions leads to force overshoot and oscillatory behavior. Explicit time-aligned scheduling mitigates these effects by ensuring that executed actions remain temporally consistent with the robot’s physical state, without reverting to conservative, blocking behavior. These results demonstrate that latency-aware execution is a critical system-level mechanism for maintaining stable interaction on industrial robotic arms with buffered control interfaces.

Although the evaluation uses a simple nonparametric policy and a single task, the observed effects are not specific to k-NN inference. Many learning-based visuomotor policies output finite-horizon action sequences, and similar timing issues arise whenever inference latency is non-negligible relative to execution dynamics. Addressing these issues at the execution interface decouples policy design from hardware-specific timing constraints, which is particularly important for manufacturing systems based on industrial robotic platforms.

%========================================================================
%                             CONCLUSION
%========================================================================
\section{Conclusion}\label{conclusion}

We presented a latency-aware framework for deploying visuomotor policy learning on industrial robotic arms. By explicitly addressing the observation–execution gap induced by sensing, inference, and execution latency, the framework targets a core limitation that has constrained the transfer of recent advances in visuomotor policy learning to industrial robotic arms. By integrating multimodal sensing, low-latency communication, latency-calibrated data handling, and VR-based teleoperation for expert demonstration collection, the framework provides the infrastructure needed to execute latency-calibrated policies on industrial hardware.

An experimental evaluation of a contact-rich assembly task demonstrates that explicit latency compensation is critical for stable and efficient policy execution on industrial robotic arms. Compared to blocking and naive asynchronous execution strategies, the proposed latency-aware strategy reduces idle time while preserving smooth motion and compliant contact behavior that closely matches expert demonstrations. In contrast, naive asynchronous execution leads to unstable force interactions and oscillatory behavior during contact. \rev{Across inference latencies from 100 ms to 500 ms, latency-aware execution maintained stable task duration, motion quality, and contact behavior, whereas blocking execution exhibited increasing idle behavior and naive asynchronous execution produced substantially larger force overshoots. Quantitative evaluation further showed that latency-aware execution consistently achieved behavior closest to the demonstration reference across all latency conditions. These findings establish latency-aware execution as a necessary system-level consideration for deploying learning-based controllers on industrial robotic arms.}

\subsection{Limitations and future work}

\rev{The proposed framework was implemented and experimentally validated on ABB industrial robotic arms accessed through safety-certified, high-level control interfaces for a contact-rich assembly task using proprioceptive, force/torque, and visual sensing. Accordingly, the findings should be interpreted within the scope of the robotic platform, task setup, sensing configuration, and policy instantiations evaluated in this work. While the core principles of latency calibration, multimodal synchronization, and latency-aware execution may transfer to other industrial robotic systems operating under non-negligible observation--execution gaps, broader applicability would require system-specific latency characterization, interface adaptation, and experimental validation.}

\rev{The current implementation also relies on offline latency calibration and assumes that observation and execution delays remain approximately constant over the task duration. This assumption is reasonable for the controlled experimental setup used in this study, but may not hold under time-varying communication, sensing, or controller delays. Future work should therefore investigate online latency calibration and adaptive compensation mechanisms that update delay estimates during deployment.}

\rev{An important direction for future research is the integration and evaluation of high-latency state-of-the-art generalist models, such as VLAs~\cite{black2026}, on industrial robotic arms. These models introduce substantially higher inference latency and more complex execution requirements, making explicit handling of the observation--execution gap even more critical. Evaluating how latency-aware execution strategies scale to such models on industrial hardware, including execution strategies for high-latency models~\cite{gr2025b, nvidia2025}, represents a promising extension of this work.}

\rev{Additional future work should examine how observation, inference, and execution latencies accumulate across multi-stage, long-horizon assembly workflows and how such latencies should be modeled during simulation-based policy training to improve temporal consistency and sim-to-real transfer on physical industrial robotic systems.}

%========================================================================

\section*{Acknowledgments}
This research was supported by the Princeton Catalysis Initiative grant, National Science Foundation (NSF, Award No. 2122271), and the School of Architecture at Princeton University. 

\section*{Data availability statement}
The data used in this study are available upon request. 

\bibliographystyle{IEEEtran} 
\bibliography{bibliography.bib}

\end{document}